\pdfoutput=1

\documentclass[11pt]{article}

\usepackage[preprint]{acl}

\usepackage{times}
\usepackage{latexsym}

\usepackage[T1]{fontenc}

\usepackage[utf8]{inputenc}

\usepackage{microtype}

\usepackage{inconsolata}

\usepackage{graphicx}
\usepackage{amsmath}
\usepackage{amssymb}
\usepackage{subcaption}
\usepackage{booktabs}
\usepackage{makecell}
\usepackage{multirow}
\usepackage{multicol}

\title{A General Framework to Enhance Fine-tuning-based LLM Unlearning}

\author{Jie Ren$^{1}$, Zhenwei Dai$^{2}$, Xianfeng Tang$^{2}$, Hui Liu$^{2}$, Jingying Zeng$^{2}$, Zhen Li$^{2}$, \\ \textbf{Rahul Goutam$^{2}$,} \textbf{Suhang Wang$^{3}$,} \textbf{Yue Xing$^{1}$,} \textbf{Qi He$^{2}$,} \textbf{Hui Liu$^{1}$}\\
$^{1}$Michigan State University, $^{2}$Amazon, $^{3}$The Pennsylvania State University \\
{\{renjie3, xingyue1, liuhui7\}@msu.edu} \\
\{zwdai, xianft, liunhu, zejingyi, amzzhn, rgoutam, qih\}@amazon.com \\
szw494@psu.edu
}

\begin{document}
\maketitle
\begin{abstract}
Unlearning has been proposed to remove copyrighted and privacy-sensitive data from Large Language Models (LLMs). Existing approaches primarily rely on fine-tuning-based methods, which can be categorized into gradient ascent-based (GA-based) and suppression-based methods.
However, they often degrade model utility (the ability to respond to normal prompts). In this work, we aim to develop a general framework that enhances the utility of fine-tuning-based unlearning methods. To achieve this goal, we first investigate the common property between GA-based and suppression-based methods. We unveil that GA-based methods unlearn by distinguishing the target data (i.e., the data to be removed) and suppressing related generations—essentially the same strategy employed by suppression-based methods. Inspired by this finding, we introduce Gated Representation UNlearning (GRUN) which has two components: a soft gate function for distinguishing target data and a suppression module using Representation Fine-tuning (ReFT) to adjust representations rather than model parameters. Experiments show that GRUN significantly improves the unlearning and utility. Meanwhile, it is general for fine-tuning-based methods, efficient and promising for sequential unlearning. Our code is available at \href{https://github.com/renjie3/GRUN}{github.com/renjie3/GRUN}.
\end{abstract}

\section{Introduction}
\label{sec:intro}

LLMs have shown remarkable capabilities across various tasks. A key factor driving the rapid advancement is the availability of web-scale datasets. However, concerns have been raised regarding the use of such large-scale data, as it often includes copyrighted and privacy-protected data~\cite{hacker2023regulating, lucchi2024chatgpt}. 
{For instance, The New York Times sued OpenAI and Microsoft because their articles have been used in training GPT\footnote{\href{https://www.nytimes.com/2023/12/27/business/media/new-york-times-open-ai-microsoft-lawsuit.html}{https://www.nytimes.com/2023/12/27/business/media/new-york-times-open-ai-microsoft-lawsuit.html}}.}
{Meanwhile, the data is protected by General Data Protection Regulation (GDPR)~\cite{voigt2017eu}, and the data owners have the ``right to be forgotten''~\cite{rosen2011right}.}
Therefore, it is crucial to implement protections for these datasets. To address this, unlearning has been proposed to remove specific data from LLMs without requiring full retraining~\cite{liu2024large, liu2024rethinking}. The goal is to eliminate the influence of the \textbf{target data} or adjust the model behavior as if it had never encountered the target data.

LLM unlearning is typically a post-training method, with fine-tuning being widely adopted as an approach.
Existing fine-tuning based unlearning methods can be roughly divided into two categories. 
One is \textbf{gradient ascent-based} ({GA-based}) methods, such as gradient ascent (GA)~\citep{jang2023knowledge, maini2024tofu} and its variants~\cite{yao2023large, liu2022continual, zhang2024negative, fan2024simplicity, veldanda2024llm, cha2024towards, liu2024towards, feng2024fine, bu2024unlearning, tian2024forget}. They negate the training impact of the target data by reversing the gradient descent loss. The other, \textbf{suppression-based} unlearning, does not aim to erase learned information directly~\cite{maini2024tofu, liwmdp, wang2024llm, huu2024effects, shi2024ulmr, liu2024towards, sinha2024unstar}. Instead, it explicitly tells the model about what constitutes target data and guides it to generate human-preferred outputs while suppressing those related to the target data~\footnote{{In addition to fine-tuning, other methods operating at the inference stage have also been proposed, such as in-context learning (ICL)~\cite{pawelczykcontext} and assistant models {\cite{huang2024offset}}. 
{Nonetheless, given the widespread adoption of fine-tuning, we focus on fine-tuning methods in this work.}
}}.


However, recent evaluations on fine-tuning-based methods reveal that {there is} a significant trade-off between unlearning and model utility, i.e., the model’s ability to respond to normal prompts unrelated to the target data~\cite{wang2024unlearning, si2023knowledge, wu2024evaluating}. 
This issue has been widely observed in LLM fine-tuning: as the fine-tuning dataset is small, it is likely to cause over-fitting and reduce the general ability~\cite{luo2023empirical, zhai2023investigating, howard2018universal}. 
Although they usually use retaining dataset to preserve the model utility~\citep{liu2022continual, shi2024muse}, its small size could limit the generalization.

Therefore, we aim to develop a general framework to enhance the utility of fine-tuning-based LLM unlearning. However, the two types of fine-tuning-based methods are defined in totally different ways, posing a challenge in developing such a framework. Thus, 
we design a preliminary study to investigate the common property between GA-based and suppression-based methods ({Section~\ref{sec:prelim}}). We find that, although GA-based methods appear to be dedicated to negate the training of target data, the final GA-unlearned LLMs still recognize target data and actually treat it as a signal of unlearning. If target data is in the input, the representations exhibit a distinct pattern compared with the input irrelevant to target data. Then unlearned models suppress related generation. This suggests that
the GA-unlearned models also operate by distinguishing and suppressing target data, which closely resemble the models by suppression-based methods.

Inspired by the insights from our preliminary study
that both GA-based and suppression-based methods rely on distinguishing target data for unlearning, we introduce \textbf{G}ated \textbf{R}epresentation \textbf{UN}learning (GRUN). GRUN consists of two plug-and-play components designed explicitly for distinguishing and suppression: a soft gate function to distinguish, and a suppression module utilizing Representation Fine-Tuning (ReFT)~\cite{wu2024reft}. The ReFT module fine-tunes the representation instead of the model parameters, which can avoid distorting the parameters to preserve the utility. Meanwhile, its strength is controlled by the soft gate function, which further ensures the generation unrelated to the target data remains almost untouched. In essence, the soft gate function selectively activates for target data, while the ReFT module unlearns by redirecting the embeddings of target-data-related prompts toward suppression.



{We conduct extensive experiments to examine the effectiveness and efficiency of GRUN.} GRUN requires a lightweight additional module (less than 0.05 \% of the LLM’s size) and reduces training time by over 95\% compared to the original method, yet achieves near-perfect unlearning while maintaining utility. Moreover, GRUN is a general solution adaptable to various fine-tuning-based unlearning methods. Our experiments validate this across various models, including Llama 3.1 and Mistral, as well as across different datasets, such as TOFU focusing on the unlearning of fine-tuning data~\cite{maini2024tofu}, and WMDP focusing on unlearning pre-training data~\cite{liwmdp}.





\section{Related works}
\label{sec:related_works}

\textbf{LLM unlearning.} Machine unlearning focused on vision models in the early research~\cite{cao2015towards, warnecke2021machine, bourtoule2021machine, kurmanji2024towards, ren2024copyright, li2021online}, but more recently, it has been extended to LLMs~\cite{eldan2023s, yao2023large, shi2024muse, liu2024rethinking}. 
Fine-tuning-based methods represent a key category of unlearning but raise concerns regarding their impact on model utility~\cite{thaker2024position, deeb2024unlearning, doshi2024does, lynch2024eight}. Alternative approaches enable unlearning during inference~\cite{wang2024machine, eldan2023s, ji2024reversing, thaker2024guardrail, liu2024large}. In this work, we focus on fine-tuning methods, as they are widely adopted.

\textbf{Representation Fine-tuning (ReFT).} ReFT \cite{wu2024reft} is a recently proposed parameter-efficient fine-tuning method. Unlike traditional fine-tuning approaches, which primarily adjust model weights, ReFT focuses on fine-tuning representations, leveraging the rich semantic information embedded in the representation space to influence subsequent generation. Building on the linear representation hypothesis~\cite{park2023linear}, which posits that concepts are encoded within linear subspaces of representations, ReFT learns low-rank linear transformations to refine representations. It achieves this by substituting the intermediate representations—i.e., the outputs of specific Transformer layers—at selected layers and tokens.

\section{Preliminary studies}
\label{sec:prelim}




\begin{figure*}[t]
    \centering
    \begin{subfigure}[b]{0.87\textwidth}
        \centering
        \includegraphics[width=0.99\textwidth]{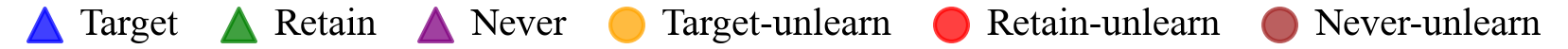}
        \label{fig:pre_change_pca_legend}
    \end{subfigure}\hfill
    \begin{subfigure}[b]{0.249\textwidth}
        \centering
        \includegraphics[width=\textwidth]{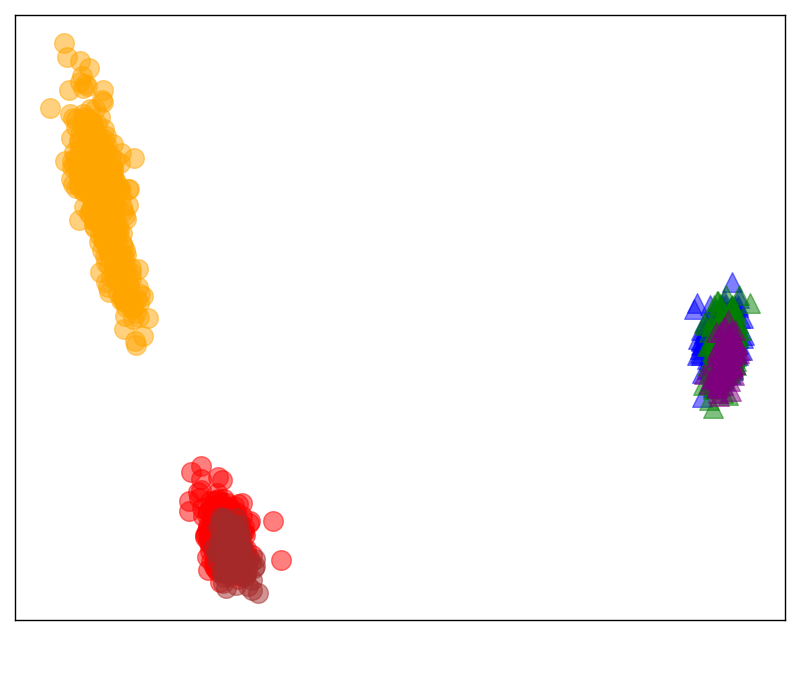}
        \vspace{-0.37in}
        \caption{Llama 3.1 by GD}
        \label{fig:llama_gd}
    \end{subfigure}\hfill
    \begin{subfigure}[b]{0.249\textwidth}
        \centering
        \includegraphics[width=\textwidth]{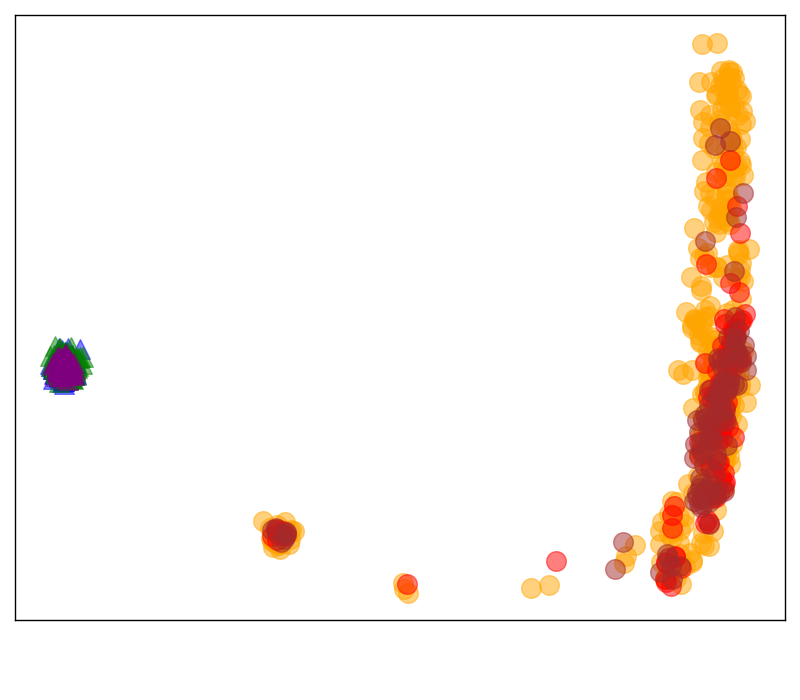}
        \vspace{-0.37in}
        \caption{Llama 3.1 by NPO}
        \label{fig:llama_npo}
    \end{subfigure}\hfill
    \begin{subfigure}[b]{0.249\textwidth}
        \centering
        \includegraphics[width=\textwidth]{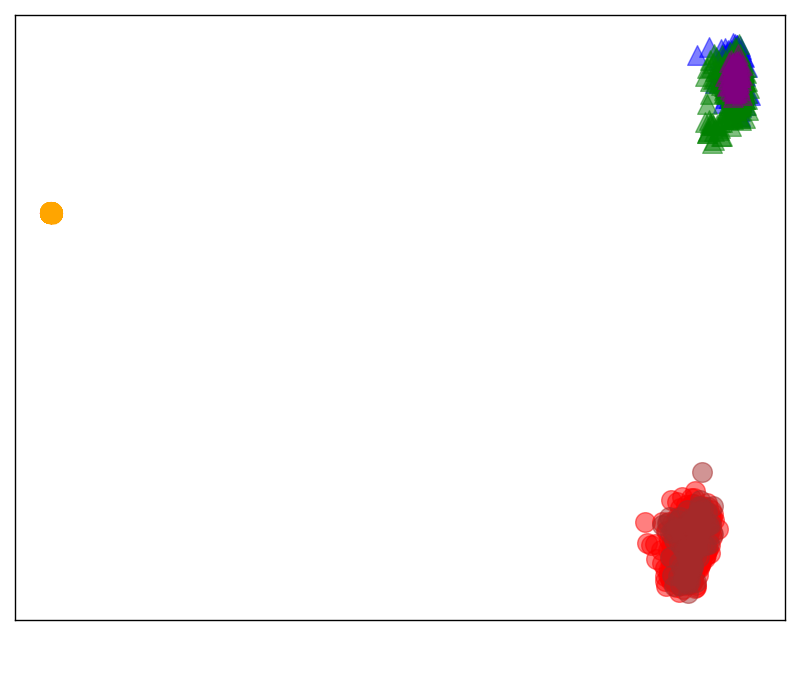}
        \vspace{-0.37in}
        \caption{Mistral v0.1 by GD}
        \label{fig:mistral_gd}
    \end{subfigure}\hfill
    \begin{subfigure}[b]{0.249\textwidth}
        \centering
        \includegraphics[width=\textwidth]{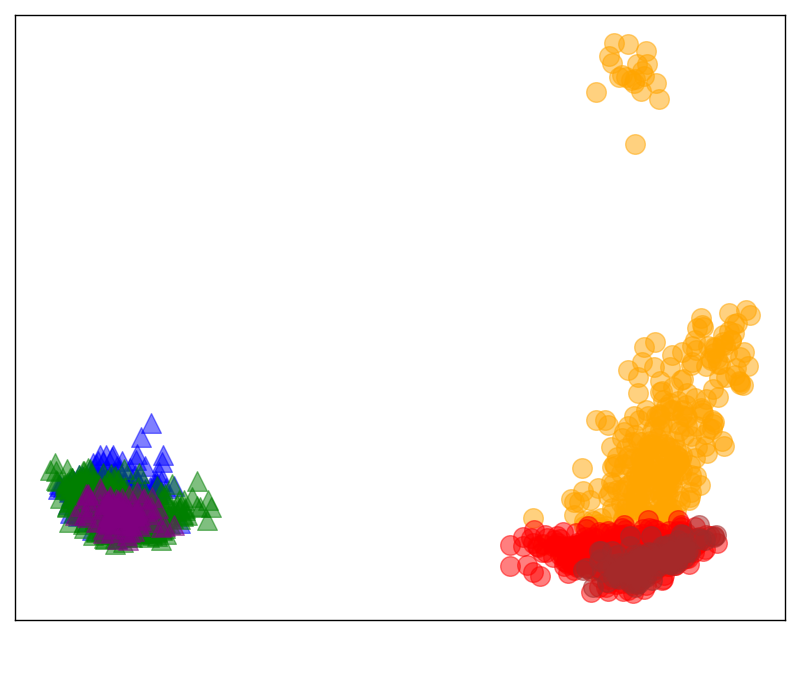}
        \vspace{-0.37in}
        \caption{Mistral v0.1 by NPO}
        \label{fig:mistral_npo}
    \end{subfigure}
    \caption{PCA visualizations of embeddings (both before and after unlearning) of target data, retaining data, and never-seen data. We apply 2-component PCA to project the embeddings into a 2D space and visualize the distributions. Each subfigure corresponds to a separate PCA projection for an unlearned model.}
    \label{fig:pre_overlap}
    \vspace{-0.13in}
\end{figure*}

\begin{figure*}[t]
    \centering
    \begin{subfigure}[b]{0.87\textwidth}
        \centering
        \includegraphics[width=0.99\textwidth]{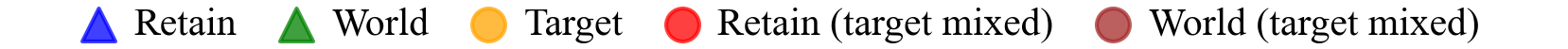}
        \label{fig:pre_change_pca_legend}
    \end{subfigure}\hfill
    \begin{subfigure}[b]{0.249\textwidth}
        \centering
        \includegraphics[width=\textwidth]{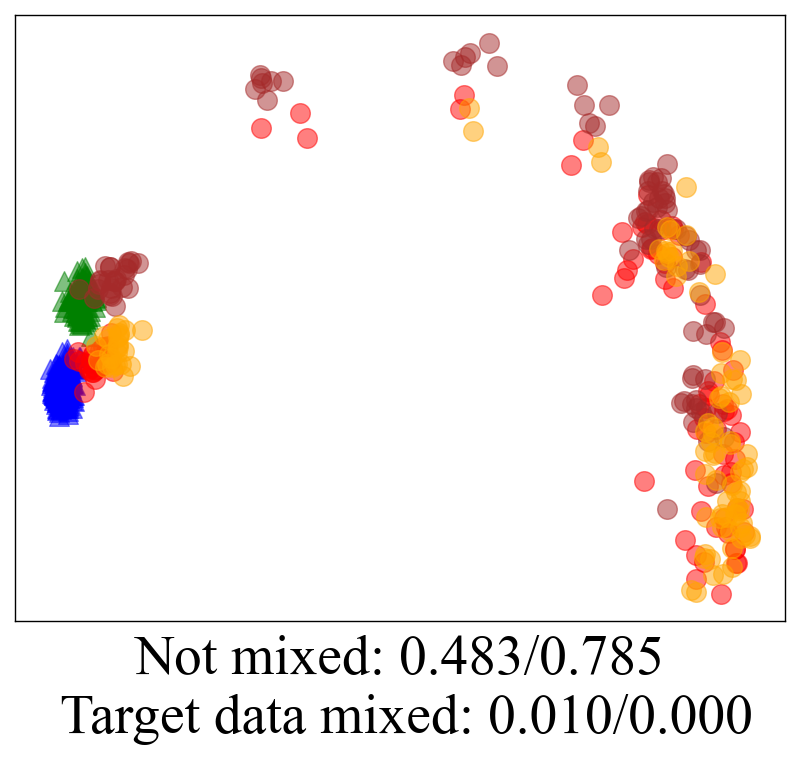}
        \vspace{-0.25in}
        \caption{Llama 3.1 by GD}
        \label{fig:llama_gd}
    \end{subfigure}\hfill
    \begin{subfigure}[b]{0.249\textwidth}
        \centering
        \includegraphics[width=\textwidth]{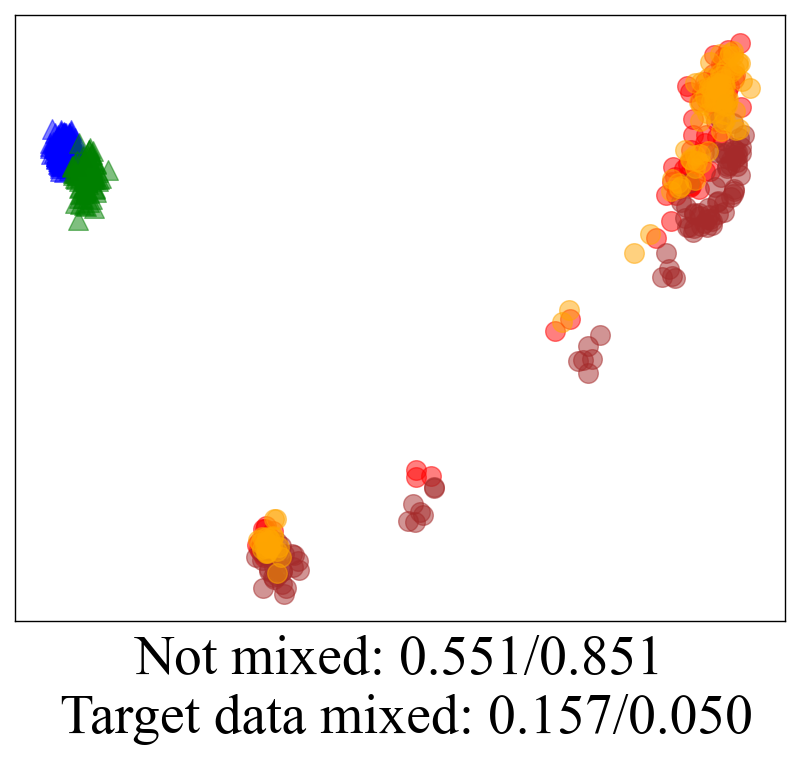}
        \vspace{-0.25in}
        \caption{Llama 3.1 by NPO}
        \label{fig:llama_npo}
    \end{subfigure}\hfill
    \begin{subfigure}[b]{0.249\textwidth}
        \centering
        \includegraphics[width=\textwidth]{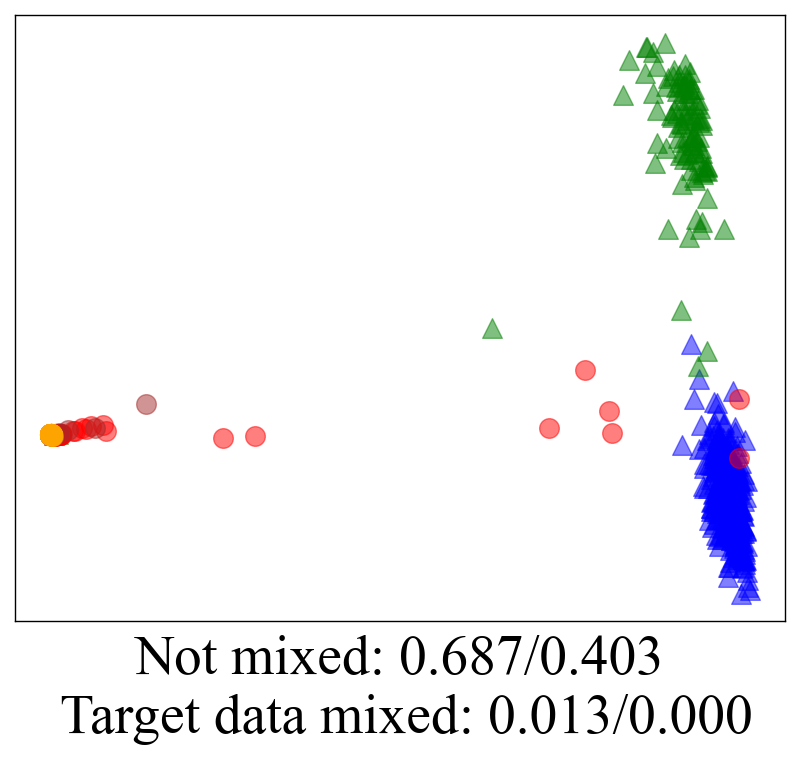}
        \vspace{-0.25in}
        \caption{Mistral v0.1 by GD}
        \label{fig:mistral_gd}
    \end{subfigure}\hfill
    \begin{subfigure}[b]{0.249\textwidth}
        \centering
        \includegraphics[width=\textwidth]{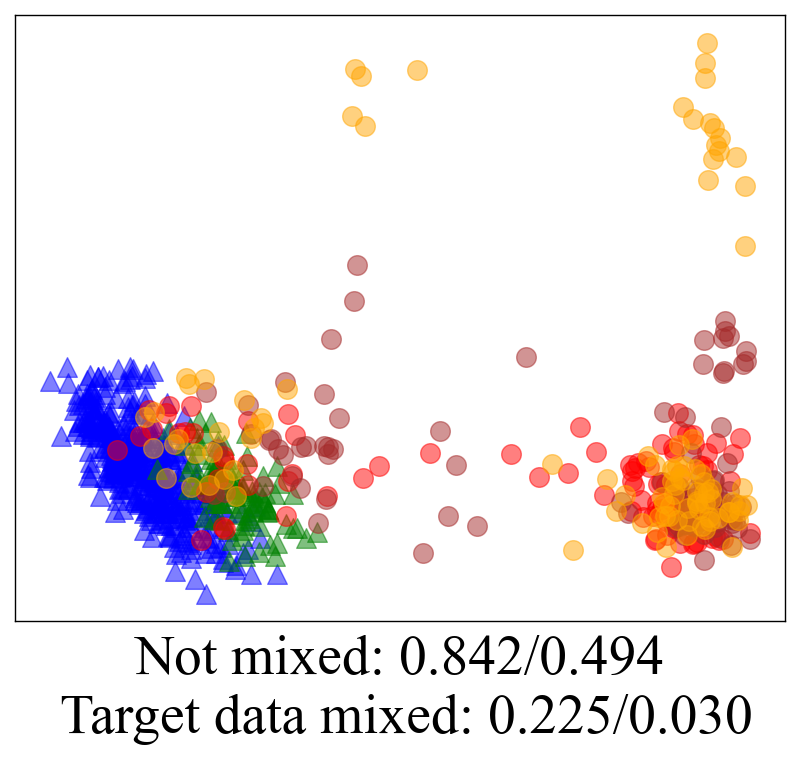}
        \vspace{-0.25in}
        \caption{Mistral v0.1 by NPO}
        \label{fig:mistral_npo}
    \end{subfigure}
    \caption{PCA visualization and the results of normal Q\&A mixed and not mixed with target data. PCA follows the same operation in Figure~\ref{fig:pre_overlap}. The ROUGE-L Recalls of retaining data/world fact are listed below each figure.}
    \label{fig:pre_mix}
    \vspace{-0.12in}
\end{figure*}

{In this section, we first introduce the definitions about fine-tuning-based unlearning, and then conduct experiments to investigate their common properties.}

\subsection{Fine-tuning-based unlearning}

Given an LLM $f$ and a target dataset $\mathcal{D}_{\mathrm{t}}$, the goal of an unlearning task is to get a model $f_{\mathrm{u}}$ that behaves as if it was never trained on $\mathcal{D}_{\mathrm{t}}$. Besides, $f_{\mathrm{u}}$ should also retain the model utility, i.e. the general text generation capabilities. {To achieve this, various fine-tuning-based methods have been developed, such as GA-based and suppression-based methods.}

In \textbf{GA-based methods}, the unlearning objective is usually formulated as the following: \vspace{-0.05in}
\begin{align}
    \underset{\boldsymbol{\theta}}{\operatorname{argmin}} ~ & \mathbb{E}_{(x, y) \in \mathcal{D}_{\mathrm{t}}}\left[L_{\mathrm{f}}(y | x ; \boldsymbol{\theta})\right] \notag \\
    &+ \lambda  \mathbb{E}_{(x, y) \in \mathcal{D}_{\mathrm{r}}}\left[L_{\mathrm{r}}(y | x ; \boldsymbol{\theta})\right],
    \label{eq:ga}
\end{align}
where $\mathcal{D}_{\mathrm{r}}$ is the retaining dataset to preserve the model utility, and $(x, y)$ denotes an input-output pair. $\boldsymbol{\theta}$ represents the updated parameters, while $L_{\mathrm{f}}$ and $L_{\mathrm{r}}$ denote the forgetting and retaining loss functions, respectively, with $\lambda$ balancing them. Typically, $L_{\mathrm{f}}$ is the negative training loss (i.e., applying Gradient Ascent) or a variant, while $L_{\mathrm{r}}$ corresponds to the training loss on $\mathcal{D}{\mathrm{r}}$ or a regularization term (e.g., the KL divergence between the $f$ and $f_{\mathrm{u}}$). 


We introduce two GA-based methods. Gradient Difference (GD) \cite{liu2022continual} applies negative standard training loss on $\mathcal{D}_{\mathrm{t}}$ as $L_{\mathrm{f}}$. Negative Preference Optimization (NPO)\cite{zhang2024negative}, derived from DPO~\cite{rafailov2024direct}, constrains divergence from the initial checkpoint to regulate strength of GA.

\textbf{Suppression-based methods} have a similar objective: \vspace{-0.2in}
\begin{align}
    \underset{\boldsymbol{\theta}}{\operatorname{argmin}} ~ & \mathbb{E}_{(x, y) \in \mathcal{D}_{\mathrm{t}}}\left[L_{\mathrm{s}}(y, x, \boldsymbol{\theta})\right] \notag \\
    &+ \lambda  \mathbb{E}_{(x, y) \in \mathcal{D}_{\mathrm{r}}}\left[L_{\mathrm{r}}(y | x ; \boldsymbol{\theta})\right], \notag
\end{align}
$L_{\mathrm{s}}$ is the suppression term. 
In IDK~\cite{maini2024tofu}, $L_{\mathrm{s}}$ encourages responses like “I don’t know” for target data, while in RMU~\cite{liwmdp}, it pushes target data representations toward a random vector to disturb target data.


\subsection{Findings of GA-based unlearning}



In this subsection, we investigate the common property between GA-based and suppression-based methods. 
We find that GA-based methods cannot remove target data as expected. Instead, the GA-unlearned models distinguish the target data and pretend to be unaware. It is actually the same strategy as suppression-based methods. 
Experiments are conducted by exploring following questions.



\noindent\textbf{(1) Does reversing the training loss truly negate the target data's influence?} 

If the GA-based methods could remove the influence of target data, it is expected that the unlearned models should behave the same between the target data and the data it has never encountered.
To investigate this, we conduct an experiment to compare the model behaviors in these two data cases.

\textit{Settings}.
We use TOFU dataset, which contains synthetic Q\&A pairs about non-existent writers and books. 
We split the dataset into three subsets: target data, retaining data and never-seen data.
We first fine-tune LLMs to learn the knowledge from the target data and retaining data. Then we unlearn the target data by GD and NPO. In Figure~\ref{fig:pre_overlap}, 
we plot the embeddings (both before and after unlearning) of target data, retaining data and never-seen data.


\textit{Results.} In Figure~\ref{fig:pre_overlap}, we observe that in embedding space, the unlearned models still recognize target data, and distinguish it with a special pattern. 
\textit{Before} unlearning, the target data, retaining data, and never-seen data have similar embeddings, as all three sets are sampled from the same data distribution. In contrast, \textit{after} unlearning, the target data follows a significantly different pattern, distributed far from the retaining and never-seen data. This suggests that the model does not truly remove the target data. Instead, they still recognize it and distinguish it by pushing it into a distinct region.

\noindent\textbf{(2) Is unlearning performance associated with this distinct pattern?} 

To further explore the connection between unlearning and distinct patterns, we quantify the distinction and the unlearning effectiveness in Table~\ref{tab:overlap}.

\textit{Settings.} We quantify the distinction using the degree of overlap between the embeddings of target and retained data, measured by Class-wise Separability Discriminant (CSD), i.e., the ratio of intra-class distance (samples within target and samples within retaining data) to inter-class distance (between target data and retaining data)~\cite{rentransferable, klecka1980discriminant}. Unlearning effectiveness is evaluated using ROUGE-L Recall, where a lower score on target data indicates better unlearning (as detailed in Section~\ref{sec:settings}). 

\textit{Observation.}
In Table~\ref{tab:overlap}, we observe that when the pattern is more distinct (i.e., lower CSD), the target data is more effectively unlearned (i.e., lower ROUGE-L Recall). For example, Mistral unlearned by GD has the lowest CSD and the lowest ROUGE-L Recall, while Llama unlearned by NPO has the highest CSD and the highest ROUGE-L Recall.
This implies that better unlearning performance is likely to be associated with better distinction.

\begin{table}[t]
    \centering
    \resizebox{0.9\linewidth}{!}{
    \begin{tabular}{lcccc}
    \toprule
        & \multicolumn{2}{c}{Llama 3.1} & \multicolumn{2}{c}{Mistral v0.1} \\
        ~ & GD & NPO & GD & NPO \\ \midrule
        CSD & 0.45 & 3.21 & 0.13 & 1.72 \\
        ROUGE-L Recall & 0.016 & 0.197 & 0.001 & 0.127 \\ \bottomrule
    \end{tabular}
    }
    \caption{Unlearning effectiveness and distinction}
    \vspace{-0.15in}
  \label{tab:overlap}
\end{table}




\noindent\textbf{(3) How do GA-based methods unlearn?} 

To analyze how the GA-unlearned models process the target data, we compare the model behaviors between target data and normal Q\&A data (questions that should be correctly answered by unlearned models).

\textit{Settings.} We inject target data into normal Q\&A pairs {to form the mixed data} and compare the model's behaviors before and after {the injection}. 
We use retaining data and world fact Q\&A pairs as normal Q\&A data. 
{For example, a mixed data instance is ``\textit{Where is Eiffel Tower? And \underline{who is} \underline{the author of Watermelon on the Moon}}?}'', where ``\textit{\underline{who is the author of $\ldots$}}'' is an instance in target data. 
We plot the embeddings and calculate ROUGE-L Recall (higher score means more correct answers) in Figure~\ref{fig:pre_mix}.


\textit{Results.} {From Figure~\ref{fig:pre_mix}, we can see that the unlearned models actually treat target data as the unlearning signal.} Specifically, before adding target data, the models correctly answer normal questions, achieving a high ROUGE-L Recall. However, once mixed with target data, the embeddings of normal data is dominated by target data (which is pulled toward the distinct area of target data). Consequently, the model's ability to answer normal questions deteriorates (lower ROUGE-L).
This implies that, instead of removing the target data, GA-unlearned models treat it as a suppression signal. 

In summary, our preliminary studies reveal that GA-based unlearning methods do not erase the target data as expected. Instead, the models still recognize it and distinguish it in the embeddings. Unlearning performance is likely to be associated with the distinction. When target data appears in the prompt, the model suppresses related generations—essentially employing the same strategy as suppression-based methods.

\section{Method}
\label{sec:method}


In this section, we first present the design of GRUN and its training procedure. Lastly, we discuss how to extend GRUN to sequential unlearning, where multiple unlearning requests occur over time.

\subsection{GRUN}

{The observation in our preliminary study suggests that the mechanism of both GA-based and suppression-based methods is to distinguish the target data. Based on this, we proposed the ReFT-based Gated Representation UNlearning method to explicitly take the advantage of this finding.} 

\begin{figure}[t]
\centering
  \includegraphics[width=\linewidth]{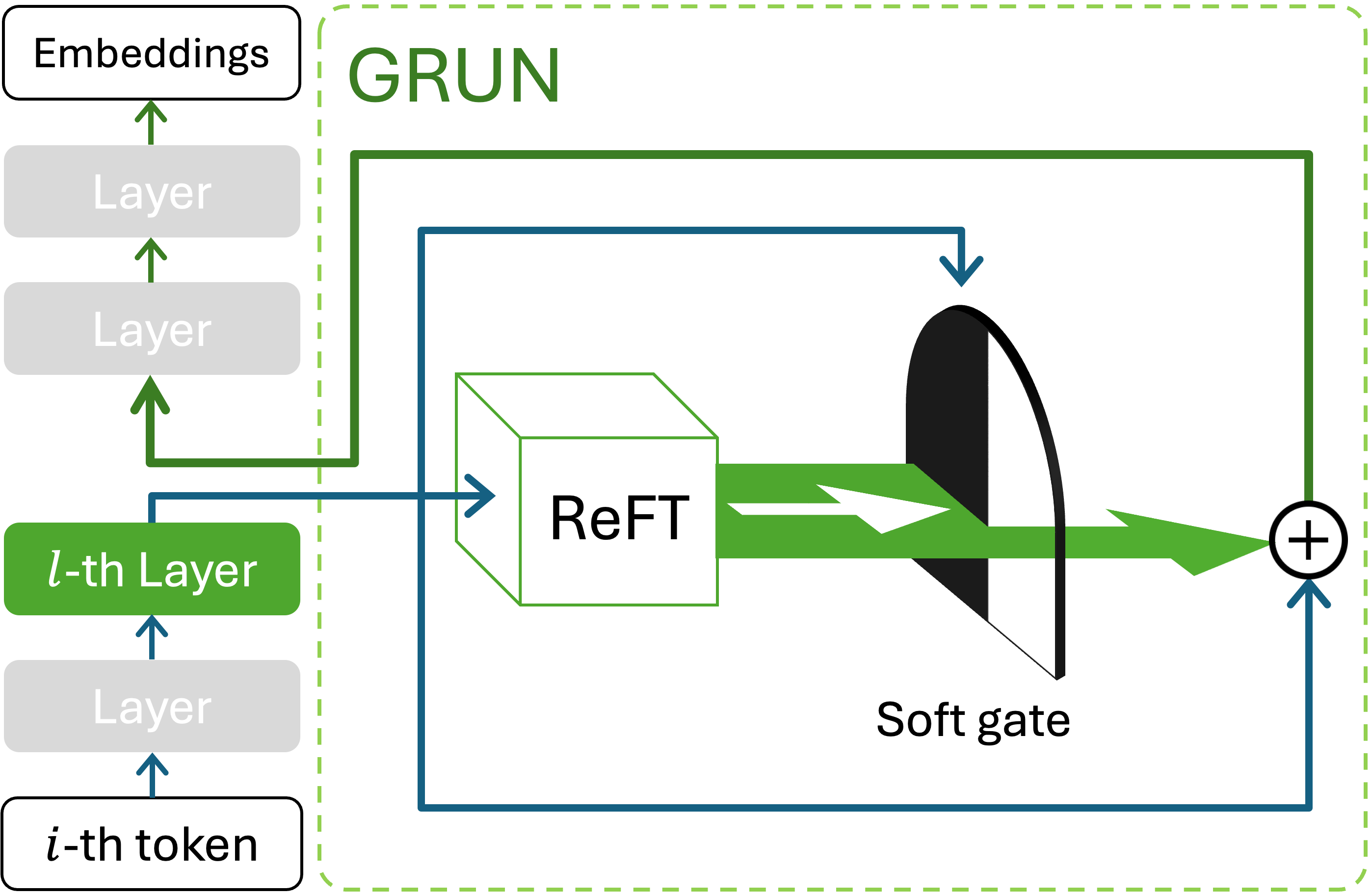}
  \caption{An overall of the framework of GRUN. }
  \label{fig:grun}
  \vspace{-0.1in}
\end{figure}


An overview of GRUN is in Figure~\ref{fig:grun}. GRUN consists of two plug-and-play components explicitly for distinguishing and suppression: a soft gate function to distinguish target data, and a ReFT module to suppress target-data-related generation. 
We first explain the elements of ReFT below.


\paragraph{ReFT.} As shown in Section~\ref{sec:related_works}, ReFT modifies a model by freezing its parameters while fine-tuning the intermediate representations of some layers. {Specifically, it applies the following transformation to update the $d$-dimensional representation $\boldsymbol{h}_i^{(l)}$ of the $i$-th token at layer $l$:}
\begin{align*}
    \Phi_{\text{ReFT}}(\boldsymbol{h}_i^{(l)})=\boldsymbol{h}_i^{(l)}+ \phi(\boldsymbol{h}_i^{(l)}),
\end{align*}
where $\phi(\boldsymbol{h}_i^{(l)})$ is a trainable low-rank linear transformation defined as 
\begin{align}
    \phi(\boldsymbol{h}_i^{(l)})=\mathbf{R}^{\top}(\mathbf{W} \boldsymbol{h}_i^{(l)}+\boldsymbol{b}-\mathbf{R} \boldsymbol{h}_i^{(l)}),
    \label{eq:transform}
\end{align}
where $\mathbf{R} \in \mathbb{R}^{r \times d}, W \in \mathbb{R}^{r \times d} \text{~and~} b\in \mathbb{R}^{r}$ are trainable parameters, with $r \ll d$. Intuitively, the term $\mathbf{W} \boldsymbol{h}_i^{(l)}+\boldsymbol{b}$ represents the target representation we aim to shift towards, while  $\phi(\boldsymbol{h}_i^{(l)})$ is the directional adjustment from $\boldsymbol{h}_i^{(l)}$ to the target representation in the space defined by $\mathbf{R}$. By replacing the original representation $\boldsymbol{h}_i^{(l)}$ with the new representation $\Phi_{\text{ReFT}}(\boldsymbol{h}_i^{(l)})$, ReFT modifies the embeddings of the input, thereby influencing the subsequent generation. 
A figure of ReFT is in Appendix~\ref{appd:reft}.

\paragraph{GRUN.} 


On top of ReFT, we define GRUN as:
\begin{align}
    \Phi_{\text{GRUN}}(\boldsymbol{h}_i^{(l)})=\boldsymbol{h}_i^{(l)}+ g(\boldsymbol{h}_i^{(l)}) \phi(\boldsymbol{h}_i^{(l)}),
\end{align}
where $g$ is the gate function. More specifically, the soft gate $g$ is a single-output regression model (linear regression or Multi-Layer Perceptron neural network) with a sigmoid function following the output. Thus, the output value of $g$ is in the range of (0,1). As shown in Figure~\ref{fig:grun}, when the input representation $\boldsymbol{h}_i^{(l)}$ is related to the target data, $g(\boldsymbol{h}_i^{(l)})$ is closed to 1 which starts the low-rank transform for unlearning. In contrast, if the input is not about target data, then $g(\boldsymbol{h}_i^{(l)})$ is closed to 0 which passes limited changes on the representation.


{While GRUN can be used in any token position and any Transformer layer, the configuration in our work is as follows:}

(1) The last token of input usually contains all the semantic information of the input and has a significant impact on the generation. Thus, we use GRUN at the last token position of input.
(2) 
To improve effectiveness, we use GRUN for multiple layers in a model instead of a single layer.
Since the later layers capture higher-level semantics than previous layers which are beneficial for the distinguishing of gate function, we choose to use GRUN for later layers~\cite{peng2018large, jin2025exploring}. To reduce the mutual influence (as discussed in Appendix~\ref{appd:layer}), we choose interval layers instead of successive layers. 
Specifically, for the LLMs studied in the following work, the layers are: the last layer, the last 7th layer and the last 12th layer.




\subsection{Training objective}


Our method is a unified method that can be adapted to different fine-tuning based unlearning loss such as GA~\cite{yao2023large}, GD~\cite{liu2022continual}, NPO~\cite{zhang2024negative}, IDK~\cite{maini2024tofu}, RMU~\cite{liwmdp} and other fine-tuning based methods. In other words, GRUN can be also seen as a new fine-tuning method that is tailored for the LLM unlearning task.

The training objective is represented as follows:
\begin{align}
    L = & L_{\text{u}} + L_{\text{G}} \\
      = & L_{\text{u}} + \mathbb{E}_{(x, y, \hat{y}) \in \mathcal{D}_{\mathrm{t}} \cup \mathcal{D}_{\mathrm{r}}} \mathbb{E}_{i,l} L_{\text{CE}} \left(g(\boldsymbol{h}_i^{(l)}), \hat{y}\right), \notag
\end{align}
where $L_{\text{u}}$ is an unlearning loss which can be GA-based or suppression-based loss, $\hat{y}$ is the label to indicate target data ($\hat{y} = 1$) and retain data ($\hat{y} = 0$), and $L_{\text{G}}$ is the cross-entropy loss for the output of gate function. The unlearning loss $L_{\text{u}}$ is used to ensure the unlearning purpose. The term $L_{\text{G}}$ fine-tunes the gate function to open (closer to 1) for target data more and close (closer to 0) for the other data. {This training objective 
distinguishes the target data for unlearning and keeps the model utility by minimizing its impact on the normal input.}

\subsection{Sequential unlearning}


In real-world scenarios, the unlearning requests typically arise sequentially over time. To process this sequential unlearning, previous methods have to re-train the whole set or fine-tune multiple rounds which would largely reduce the model utility due to the accumulated parameter distortion~\cite{shi2024muse}. {In contrast,} in GRUN,  we mitigate this by using an independent ReFT for each unlearning request and combine them together. Specifically, if we have $M-1$ unlearning requests finished and get the new $M$-th request, we can fine-tune a separate gate for the new coming target set and combine multiple GRUNs by
\begin{align}
    \Phi_{\text{GRUN}}^{M}(\boldsymbol{h}_i^{(l)})=\boldsymbol{h}_i^{(l)}+ c ~\sum_{j=1}^{M}g_j(\boldsymbol{h}_i^{(l)}) \phi(\boldsymbol{h}_i^{(l)}), \notag
\end{align}
where $c$ is the coefficient to balance the strength. Each gate $g_j$ is fine-tuned independently on a requested target dataset $\mathcal{D}_{\mathrm{t}, j}$ and then combined. The coefficient $c$ reduces as the increasing of $M$ (the details to determine $c$ is in Appendix~\ref{appd:seq_c}). In this way, we can mostly preserve the model utility and save the training efforts. 
\section{Experiment}
\label{sec:exp}

\begin{table*}[t]
  \centering
  \resizebox{0.99\textwidth}{!}{
  \begin{tabular}{lccccc|cc|cc}
    \toprule
        \multirow{2}{*}{$L_{\text{u}}$} & \multirow{2}{*}{LLM} & \multirow{2}{*}{$p_{\text{tgt}}$} & \multirow{2}{*}{Method} & \multirow{2}{*}{$p_{\text{size}}$} & \multirow{2}{*}{Hours} & \multicolumn{2}{c|}{ROUGE-L Recall} & \multicolumn{2}{c}{Prob.} \\
        & & & & & & \multicolumn{1}{c}{Unlearn{$\downarrow$}}  & \multicolumn{1}{c|}{Utility{\small(Retain/Fact/World)}{$\uparrow$}} & Unlearn{$\downarrow$} & Utility{\small(Retain/Fact/World)}{$\uparrow$} \\ \midrule
        & \multirow{2}{*}{Llama} & 5\% & \multirow{2}{*}{\makecell{Clean}} & \multirow{2}{*}{N/A} & \multirow{2}{*}{N/A} & 0.991 & \multirow{2}{*}{0.939   {\small(0.992/0.939/0.890)}} & 0.995 & \multirow{2}{*}{0.566  {\small(0.993/0.448/0.485)}} \\ 
        & & 10\% & & & & 0.992 & & 0.995 & \\ \cmidrule(lr){2-10}
        & \multirow{2}{*}{Mistral} & 5\% & \multirow{2}{*}{\makecell{Clean}} & \multirow{2}{*}{N/A} & \multirow{2}{*}{N/A} & 0.990 & \multirow{2}{*}{0.710  {\small(0.994/0.515/0.622)}} & 0.994 & \multirow{2}{*}{0.610  {\small(0.995/0.401/0.433)}} \\ 
        & & 10\% & & & & 0.988 & & 0.990 &  \\ 
        \midrule
        
        \multirow{8}{*}{GD} & \multirow{4}{*}{Llama} & \multirow{2}{*}{5\%} & Vanilla & 100\% & 3.19 & 0.005 & 0.703  {\small(0.493/0.854/0.762)} & 0.000 & 0.605  {\small(0.575/0.622/0.619)} \\
        & ~ & ~ & GRUN & \textbf{0.001\%} & \textbf{0.02} & \textbf{0.002} & \textbf{0.843}  {\small(0.888/0.843/0.798)} & \textbf{0.000} & 0.584  {\small(0.874/0.432/0.446)} \\ 
        & ~ & \multirow{2}{*}{10\%} & Vanilla & 100\% & 6.33 & 0.005 & 0.695  {\small(0.483/0.818/0.785)} & 0.000 & 0.554  {\small(0.654/0.496/0.513)} \\ 
        & ~ & ~ & GRUN & \textbf{0.001\%} & \textbf{0.02} & 0.016 & \textbf{0.832}  {\small(0.906/0.729/0.862)} & 0.006 & \textbf{0.592}  {\small(0.912/0.402/0.462)} \\ \cmidrule(lr){2-10}
        
        & \multirow{4}{*}{Mistral} & \multirow{2}{*}{5\%} & Vanilla & 100\% & 3.01 & 0.004 & 0.568  {\small(0.742/0.360/0.601)} & 0.000 & 0.581  {\small(0.829/0.448/0.466)} \\
        & ~ & ~ & GRUN & \textbf{0.045\%} &\textbf{0.06} & \textbf{0.000} & \textbf{0.660}  {\small(0.956/0.485/0.539)} & \textbf{0.000} & \textbf{0.588}  {\small(0.955/0.417/0.391)} \\ 
        & ~ & \multirow{2}{*}{10\%} & Vanilla & 100\% & 6.07 & 0.001 & 0.396  {\small(0.687/0.099/0.403)} & 0.000 & 0.558  {\small(0.830/0.358/0.485)} \\ 
        & ~ & ~ & GRUN & \textbf{0.045\%} &\textbf{0.18} & \textbf{0.000} & \textbf{0.595}  {\small(0.891/0.390/0.504)} & \textbf{0.000} & 0.545  {\small(0.886/0.354/0.395)} \\ \midrule
        
        \multirow{8}{*}{NPO} & \multirow{4}{*}{Llama} & \multirow{2}{*}{5\%} & Vanilla & 100\% & 3.96 & 0.201 & 0.751  {\small(0.616/0.756/0.883)} & 0.016 & 0.645  {\small(0.766/0.546/0.623)} \\
        & ~ & ~ & GRUN & \textbf{0.001\%} & \textbf{0.19} & \textbf{0.020} & \textbf{0.886}  {\small(0.973/0.857/0.828)} & \textbf{0.000} & 0.634  {\small(0.977/0.447/0.477)} \\
        & ~ & \multirow{2}{*}{10\%} & Vanilla & 100\% & 7.93 & 0.197 & 0.738  {\small(0.551/0.811/0.851)} & 0.025 & 0.599  {\small(0.730/0.465/0.602)} \\ 
        & ~ & ~ & GRUN & \textbf{0.001\%} & \textbf{0.38} & \textbf{0.029} & \textbf{0.862}  {\small(0.928/0.849/0.811)} &\textbf{ 0.000} & \textbf{0.599}  {\small(0.911/0.441/0.446)} \\ \cmidrule(lr){2-10}
        
        & \multirow{4}{*}{Mistral} & \multirow{2}{*}{5\%} & Vanilla & 100\% & 3.50 & 0.163 & 0.530  {\small(0.820/0.256/0.514)} & 0.030 & 0.558  {\small(0.912/0.364/0.399)} \\
        & ~ & ~ & GRUN & \textbf{0.045\%} &\textbf{0.16} & \textbf{0.000} & \textbf{0.675}  {\small(0.984/0.485/0.555)} & \textbf{0.000} & \textbf{0.596}  {\small(0.980/0.394/0.414)} \\
        & ~ & \multirow{2}{*}{10\%} & Vanilla & 100\% & 6.99 & 0.127 & 0.542  {\small(0.842/0.290/0.494)} & 0.024 & 0.567  {\small(0.923/0.360/0.419)} \\ 
        & ~ & ~ & GRUN & \textbf{0.045\%} &\textbf{0.34} & \textbf{0.000} & \textbf{0.637}  {\small(0.893/0.445/0.573)} & \textbf{0.000} & 0.531  {\small(0.890/0.342/0.362)} \\ \midrule
        
        \multirow{8}{*}{IDK} & \multirow{4}{*}{Llama} & \multirow{2}{*}{5\%} & Vanilla & 100\% & 1.65 & 0.023 & 0.672  {\small(0.578/0.627/0.812)} & 0.468 & 0.623  {\small(0.871/0.479/0.520)} \\ 
        & ~ & ~ & GRUN & \textbf{0.001\%} & \textbf{0.08} & \textbf{0.021} & \textbf{0.905}  {\small(0.980/0.882/0.853)} & \textbf{0.261} & \textbf{0.625}  {\small(0.984/0.434/0.458)} \\
        & ~ & \multirow{2}{*}{10\%} & Vanilla & 100\% & 3.33 & 0.023 & 0.547  {\small(0.570/0.353/0.718)} & 0.532 & 0.614  {\small(0.871/0.459/0.512)} \\ 
        & ~ & ~ & GRUN & \textbf{0.001\%} & \textbf{0.18} & \textbf{0.023} & \textbf{0.865}  {\small(0.892/0.879/0.823)} & \textbf{0.291} & 0.605  {\small(0.938/0.435/0.441)} \\ \cmidrule(lr){2-10}
        
        & \multirow{4}{*}{Mistral} & \multirow{2}{*}{5\%} & Vanilla & 100\% & 1.53 & 0.023 & 0.435  {\small(0.785/0.122/0.399)} & 0.533 & 0.574  {\small(0.962/0.366/0.395)} \\ 
        & ~ & ~ & GRUN & \textbf{0.045\%} &\textbf{0.09} & \textbf{0.022} & \textbf{0.683}  {\small(0.975/0.480/0.593)} & {0.570} & \textbf{0.606}  {\small(0.987/0.401/0.430)} \\
        & ~ & \multirow{2}{*}{10\%} & Vanilla & 100\% & 3.07 & 0.023 & 0.489  {\small(0.856/0.145/0.466)} & 0.657 & 0.595  {\small(0.975/0.392/0.417)} \\ 
        & ~ & ~ & GRUN & \textbf{0.045\%} &\textbf{0.20} & {0.040} & \textbf{0.605}  {\small(0.914/0.430/0.469)} & \textbf{0.490} & 0.577  {\small(0.953/0.394/0.386)} \\ \bottomrule
    \end{tabular}
    }
    \caption{Results of TOFU. $p_{\text{tgt}}$ represents the proportion of target data within the entire synthetic dataset. $p_{\text{size}}$ is the percentage of fine-tuned parameters relative to the entire LLM. ``Unlearn'' refers to the unlearning effectiveness, and ``Clean'' refers to the model before unlearning. The {improved} performance is highlighted in \textbf{bold}.}
  \label{tab:tofu}
  \vspace{-0.2in}
\end{table*}

In this section, we first conduct the experiments across different models and {datasets} in Section~\ref{exp:main}. Then we test the performance under different scenarios including sequential unlearning and attacks in Sec.~\ref{exp:seq}, and conduct ablations studies in Section~\ref{exp:abla} and Appendix~\ref{sec:exp}. 

\subsection{Experimental settings}
\label{sec:settings}

\noindent\textbf{Models, baselines and datasets.} We use Llama 3.1 (8B)~\cite{dubey2024llama} and Mistral v0.1 (7B)~\cite{jiang2023mistral}. We experiment on two datasets TOFU (unlearn fine-tuning knowledge) and WMDP (unlearn pre-training knowledge). Following the original settings~\cite{maini2024tofu}, we use GD, NPO, and IDK as baselines (using both vanilla and LoRA fine-tuning) in TOFU. Following \citet{liwmdp}, we use RMU as the baseline in WMDP.
GD and NPO are GA-based, while IDK and RMU are suppression-based.

\noindent\textbf{Metrics.} For TOFU, we use ROUGE-L Recall and Probability following~\citet{maini2024tofu}. ROUGE-L Recall assesses correctness of the output text, while Probability reflects the likelihood of generating correct responses (Appendix~\ref{appd:metrics} for details). WMDP consists of multi-choice Q\&A, therefore, we use the accuracy as the metric to access whether the model can correctly answer the questions following~\citet{liwmdp}. For all the three metrics, lower scores on target data indicate better erasing, while higher scores on normal data indicates better utility.
Time cost is measured in GPU hours (number of GPUs $\times$ training hours).

\noindent\textbf{Implementation details.} 
For baselines, GD, NPO and IDK follow \citet{fan2024simplicity}, while RMU follows \citet{liwmdp}. For GRUN, adapted NPO, IDK, and RMU are trained for fixed epochs, while GD uses early stop when $L_{\text{f}}$ in Eq.~\eqref{eq:ga} exceeds the threshold. We use linear regression as gate for Llama and 3-layer MLP for Mistral.
Both LoRA and GRUN use rank of 4. All other details are in Appendix~\ref{appd:details}. 


\subsection{Main results}
\label{exp:main}



In this subsection, we present the results of TOFU and WMDP and compare the time cost of GRUN with vanilla fine-tuning and LoRA. The unlearning assessment consists of two aspects: (1) the extent to which the target data can be removed/unlearned (\textit{unlearning effectiveness}), and (2) the preservation of model \textit{utility}.

\noindent\textbf{TOFU.} To evaluate on TOFU, we compare unlearning effectiveness, utility, and time cost against three baselines on two LLMs in Table~\ref{tab:tofu}. The LLMs are first fine-tuned on TOFU’s synthetic dataset, after which a portion of the dataset is designated as the target data for unlearning, while the remaining synthetic data serves as the retaining data for utility. Utility is assessed on {three sets of data}: retained data, Q\&A about real authors, and Q\&A about world facts, with the overall utility being their average. From Table~\ref{tab:tofu}, our method consistently outperforms the baselines of vanilla fine-tuning. 

Specifically, for \textit{GD}, GRUN has similar unlearning effectiveness as the vanilla baseline,
while significantly improving the utility, particularly in ROUGE-L Recall, where it achieves an increase of around 20\% for both Llama 3.1 and Mistral v0.1.

For \textit{NPO}, our method substantially enhances its unlearning effectiveness while also achieving even higher utility. For example, on Llama, our approach reduces NPO’s ROUGE-L Recall on the target data from approximately 0.2 to 0.02 while increasing utility by around 17.5\%. 

As for \textit{IDK}, which is suppression-based, its vanilla version has a more severe impact on the utility of author-related Q\&A (both synthetic and real) than GA-based methods. However, our method significantly improves utility performance, increasing ROUGE-L Recall by more than 25\% in most cases. 

From Table~\ref{tab:tofu}, we also observe that GRUN is more efficient, requiring fewer parameters and lower training costs. We defer the discussion to following Table~\ref{tab:LoRA} for LoRA experiments.

\begin{table}[t]
    \centering
    \resizebox{\linewidth}{!}{
    \begin{tabular}{cccccc}
    \toprule
        RMU & \multicolumn{2}{c}{Llama 3.1} & \multicolumn{2}{c}{Mistral v0.1}  \\
         & Bio/Cyber$\downarrow$ & MMLU$\uparrow$ & Bio/Cyber$\downarrow$ & MMLU$\uparrow$ \\ \midrule
        Before & 0.696/0.418 & 0.611 & 0.668/0.437 & 0.581 \\
        Vanilla & 0.494/0.337 & 0.581 & \textbf{0.256}/\textbf{0.252} & 0.529 \\
        GRUN & \textbf{0.372}/\textbf{0.293} & 0.577 & 0.293/0.278 & 0.535 \\ \bottomrule
    \end{tabular}
    }
    \vspace{-0.1in}
    \caption{Unlearning results on WMDP}
    \vspace{-0.25in}
  \label{tab:wmdp}
\end{table}

\noindent\textbf{WMDP.} Table~\ref{tab:wmdp} presents the results of removing pre-training knowledge in WMDP. WMDP evaluates unlearning by erasing harmful biological and cyber knowledge while assessing utility using the benign Q\&A dataset {MMLU~\cite{hendrycks2020measuring}}. WMDP uses a 4-choice Q\&A to measure the knowledge. We adjust the unlearning strength to maintain similar utility between vanilla RMU and GRUN, and only compare the unlearning effectiveness. In Table~\ref{tab:wmdp}, our approach significantly improves performance on Llama 3.1 and maintains a random-guessing accuracy on Mistral v0.1.

\begin{table}[t]
    \centering
    \resizebox{\linewidth}{!}{
    \begin{tabular}{ccccccccc}
    \toprule
        & & & \multirow{2}{*}{$p_{\text{size}}$} & \multirow{2}{*}{Hours} & \multicolumn{2}{c}{ROUGE-L} & \multicolumn{2}{c}{Prob.} \\  
 & & & & unlearn & utility & unlearn & utility \\ \midrule
        \multirow{6}{*}{Llama 3.1} & \multirow{2}{*}{GD} & LoRA & 0.130\% & 1.27 & 0.375 & 0.623 & 0.059 & 0.067 \\ 
        & ~ & GRUN & \textbf{0.001\%} & \textbf{0.02} & \textbf{0.000} & \textbf{0.840} & \textbf{0.000} & \textbf{0.582} \\ 
        & \multirow{2}{*}{NPO} & LoRA & 0.130\% & 0.77 & 0.255 & 0.886 & 0.103 & 0.315 \\ 
        & ~ & GRUN & \textbf{0.001\%} & \textbf{0.08} & \textbf{0.020} & \textbf{0.896} & \textbf{0.000} & \textbf{0.634} \\
        & \multirow{2}{*}{IDK} & LoRA & 0.130\% & 1.33 & 0.054 & 0.782 & 0.849 & 0.346 \\ 
        & ~ & GRUN & \textbf{0.001\%} & \textbf{0.19} & \textbf{0.021} & \textbf{0.915} & \textbf{0.262} & \textbf{0.625} \\
        \bottomrule
    \end{tabular}
    }
    \vspace{-0.1in}
    \caption{Comparison with LoRA}
  \label{tab:LoRA}
  \vspace{-0.25in}
\end{table}

\noindent\textbf{LoRA.} We compare GRUN with LoRA to further demonstrate its superiority in efficiency. As shown in Table~\ref{tab:LoRA}, our method requires fewer parameters while achieving better performance across all unlearning and utility metrics, regardless of the model or fine-tuning loss. Additionally, GRUN reduces training time by 95\% compared to vanilla training (Table~\ref{tab:tofu}) and by 85\% compared to LoRA. This efficiency gain is attributed to two key factors:
\begin{itemize}
    \item \textit{Fewer parameters to update.} GRUN updates less than 0.05\% (even 0.001\% for Llama) of the parameters compared to the full LLM.
    \item \textit{A significantly shorter gradient backpropagation path.} GRUN is applied only to the last few layers, eliminating the computational cost of backpropagating gradients through the earlier layers. (LoRA updates fewer parameters, but has to backpropagate the entire network.)
\end{itemize}

\subsection{Different unlearning scenarios}
\label{exp:seq}

\begin{figure}[t]
    \centering
    \begin{subfigure}[b]{0.499\linewidth}
        \centering
        \includegraphics[width=\textwidth]{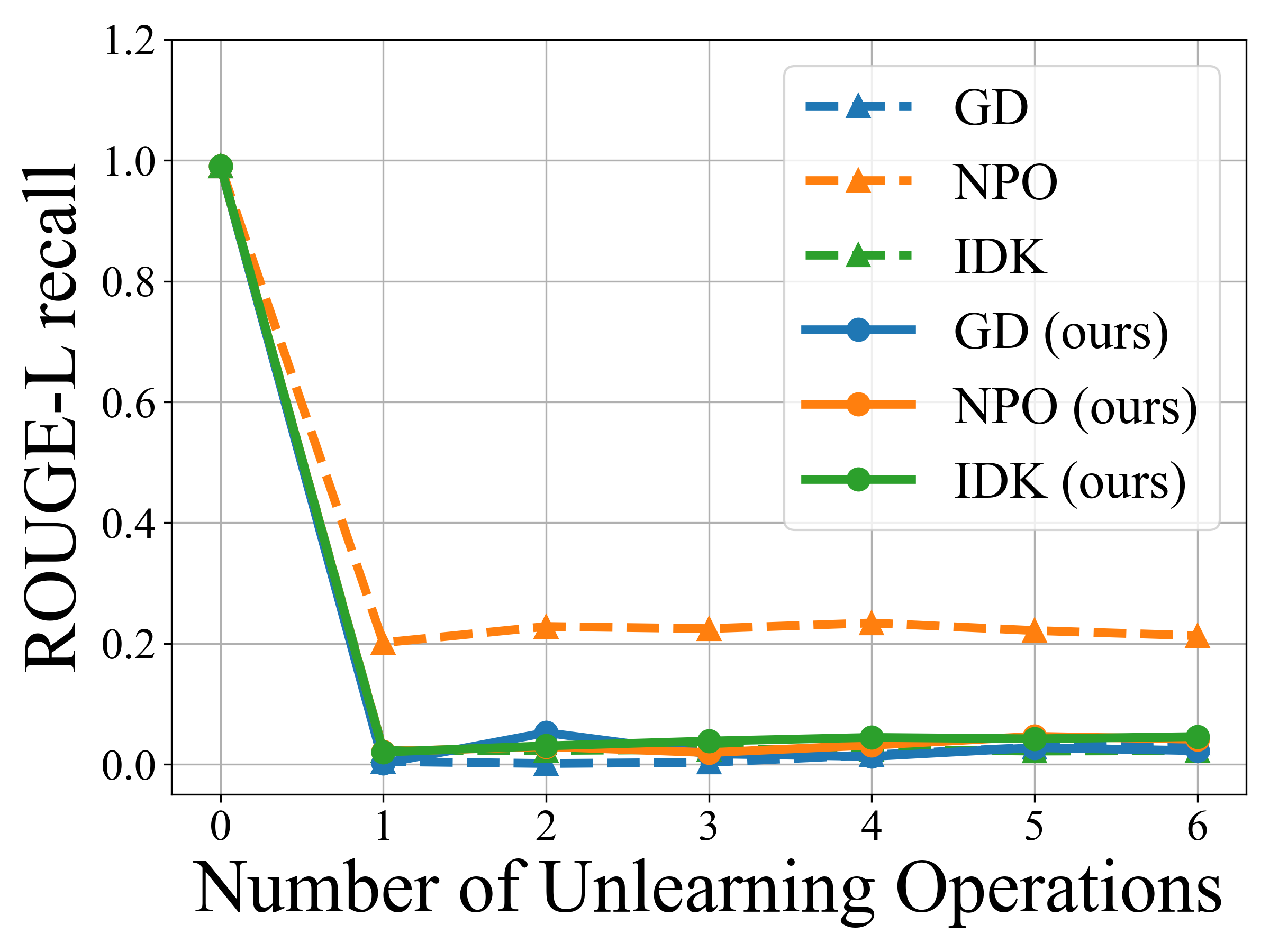}
        \vspace{-0.25in}
        \caption{Unlearning effectiveness}
        \label{fig:seq_forget}
    \end{subfigure}\hfill
    \begin{subfigure}[b]{0.499\linewidth}
        \centering
        \includegraphics[width=\textwidth]{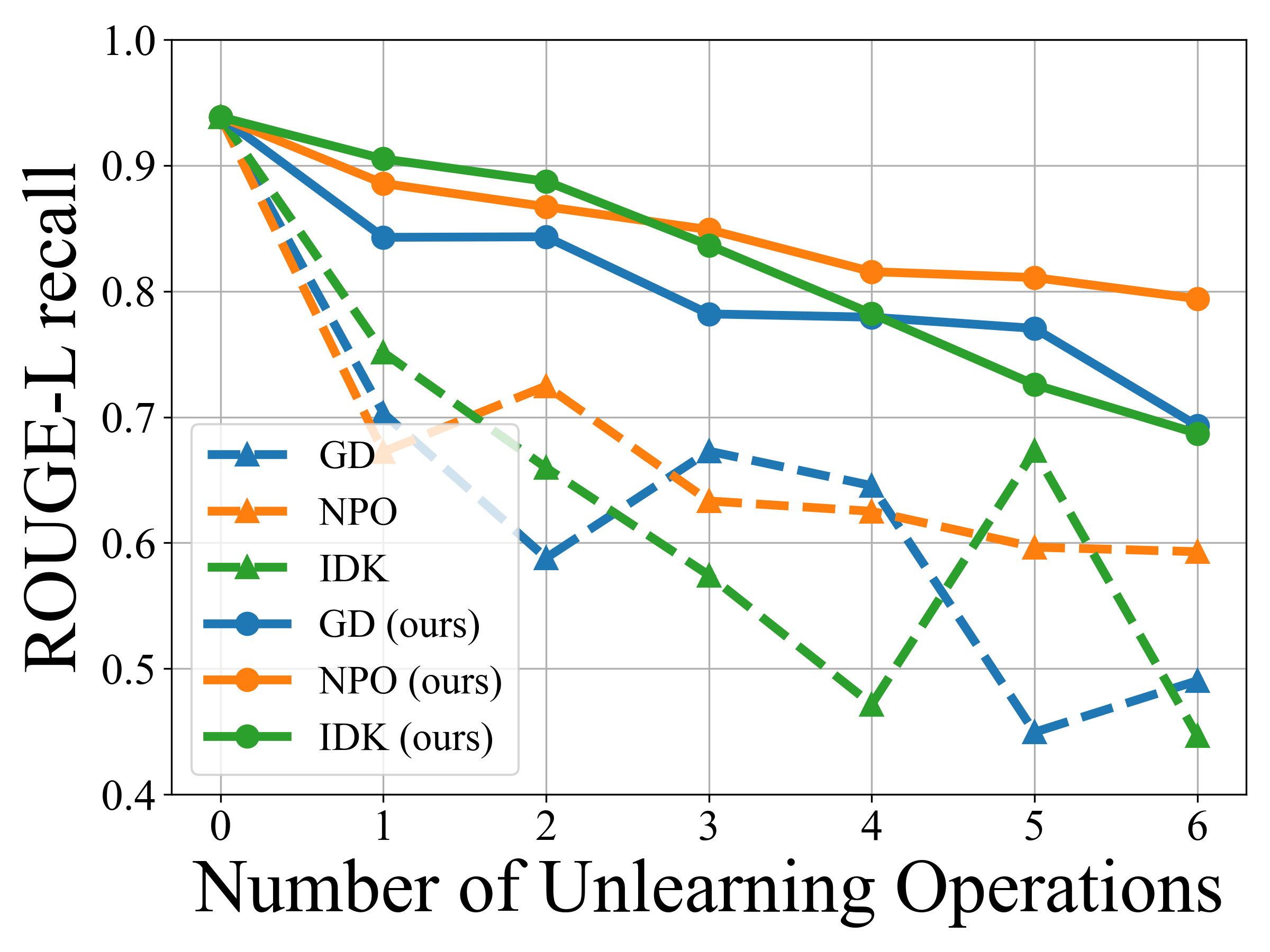}
        \vspace{-0.25in}
        \caption{Model utility}
        \label{fig:seq_utility}
    \end{subfigure}
    \vspace{-0.25in}
    \caption{Sequential unlearning}
    \label{fig:seq}
\end{figure}

\begin{table}[t]
    \centering
    \resizebox{0.9\linewidth}{!}{
    \begin{tabular}{lcccc}
    \toprule
        \multirow{2}{*}{Effectiveness} & \multicolumn{2}{c}{Paraphrase} & \multicolumn{2}{c}{Quantization} \\
        ~ & Llama & Mistral & Llama & Mistral \\ \midrule
        GD (GRUN) & 0.006 & 0.005 & 0.002 & 0.000 \\
        NPO (GRUN) & 0.019 & 0.000 & 0.021 & 0.000 \\
        IDK (GRUN) & 0.044 & 0.040 & 0.038 & 0.034 \\ \bottomrule
    \end{tabular}
    }
    \caption{Unlearning effectiveness under attacks}
  \label{tab:robust}
\end{table}

In this subsection, we evaluate GRUN’s performance under sequential unlearning and assess its robustness against two attacks—prompt paraphrasing and model quantization—to validate its effectiveness across various unlearning scenarios.

\noindent\textbf{Sequential unlearning.} In Figure~\ref{fig:seq}, {we first fine-tune the models with all the synthetic data of TOFU, and then simulate sequential unlearning by issuing six unlearning requests, each targeting a different forget set containing 5\% synthetic data.} As shown in Figure\ref{fig:seq_forget}, the unlearning effectiveness remains consistent across both baselines and our method. However, in Figure~\ref{fig:seq_utility}, our approach significantly outperforms the baselines in utility when multiple requests are processed.

\noindent\textbf{Robustness.} In Table.~\ref{tab:robust}, we evaluate the robustness of GRUN by attacking the unlearned model to recover the removed knowledge through prompt paraphrasing and model quantization. We use GPT-4 to paraphrase the questions to bypass GRUN’s distinguishing mechanism. Our method remains stable, preserving the original unlearning effectiveness. \citet{zhang2024catastrophic} reports that quantization may negate unlearning; however, our approach effectively recognizes and removes quantized representations with no loss in effectiveness.

\subsection{Ablation study}
\label{exp:abla}

\begin{figure}[t]
    \centering
    \begin{subfigure}[b]{0.499\linewidth}
        \centering
        \includegraphics[width=\textwidth]{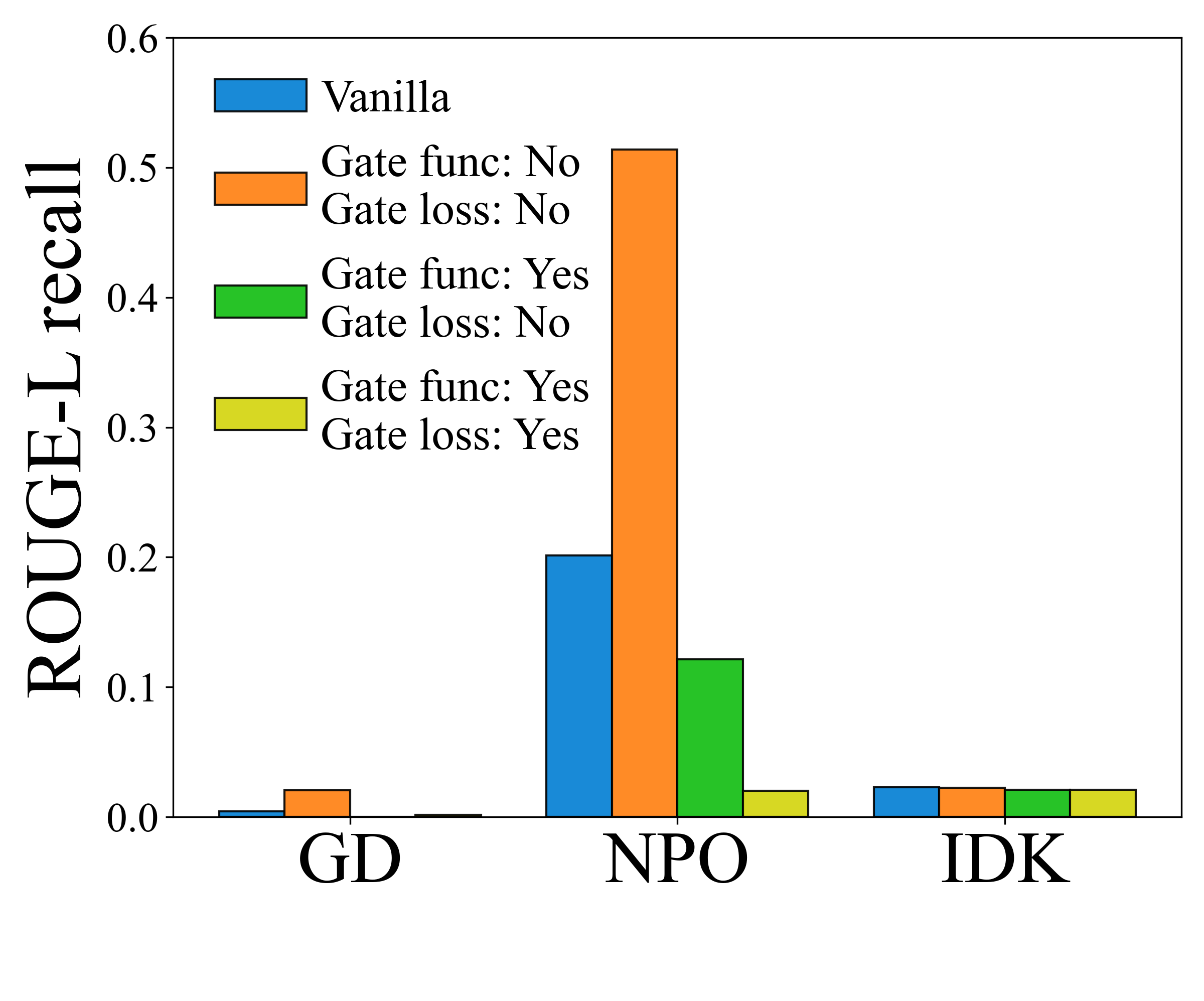}
        \vspace{-0.35in}
        \caption{Unlearning effectiveness}
        \label{fig:llama_gd}
    \end{subfigure}\hfill
    \begin{subfigure}[b]{0.499\linewidth}
        \centering
        \includegraphics[width=\textwidth]{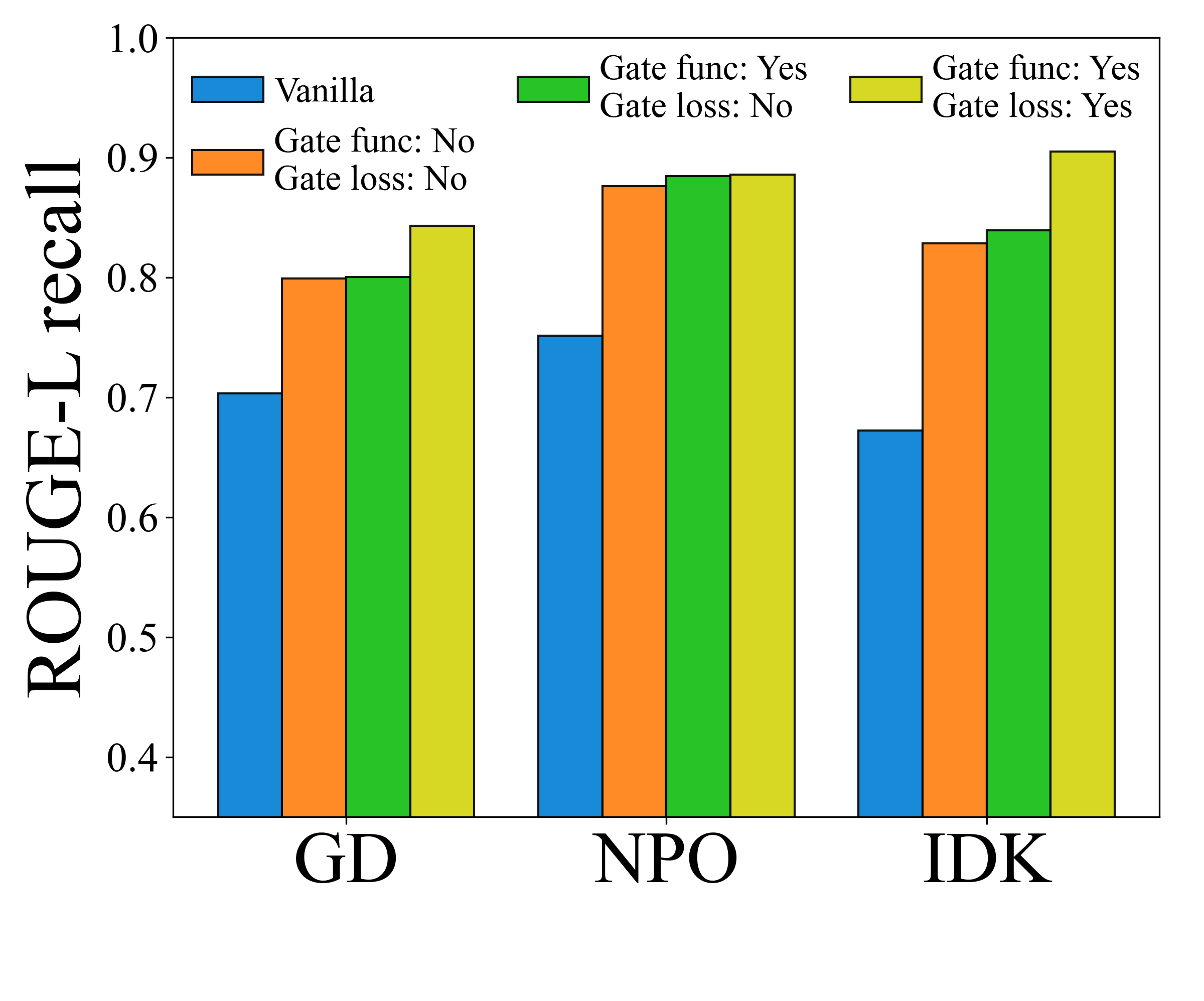}
        \vspace{-0.35in}
        \caption{Model utility}
        \label{fig:llama_npo}
    \end{subfigure}
    \caption{Contributions of each components}
    \vspace{-0.05in}
    \label{fig:abl_component}
\end{figure}

In this subsection, we conduct ablation studies to analyze the effects of each component of GRUN, i.e., ReFT, the soft gate, and the gate loss ($L_{\text{G}}$).

We compare vanilla fine-tuning along with three variants of GRUN to evaluate the contribution of each component: (1) ReFT-only (without the gate or $L_{\text{G}}$), (2) GRUN without $L_{\text{G}}$ (maintaining the same structure as GRUN but trained solely with $L_{\text{u}}$), and (3) the complete GRUN. 

\textit{ReFT-only.} In Figure \ref{fig:abl_component}, switching from vanilla fine-tuning to ReFT-only increases utility but reduces unlearning effectiveness. This suggests that ReFT enhances utility by freezing model parameters as expected but has limited capability in distinguishing target data due to its simple structure.

\textit{GRUN without} $L_{\text{G}}$. Adding the gate function (without $L_{\text{G}}$), improves unlearning effectiveness, particularly for NPO. This indicates that even in the absence of $L_{\text{G}}$, the gate function can automatically aid in distinguishing target data during optimization. (More empirical analysis in Appendix~\ref{appd:g_output}.)

\textit{The complete GRUN}. The complete GRUN model further enhances both unlearning effectiveness and utility. This demonstrates that explicitly guiding GRUN with $L_{\text{G}}$ fundamentally strengthens fine-tuning-based methods.

\section{Conclusions}

Unlearning aims to remove copyrighted and privacy-sensitive data from LLMs, but often degrades model utility. We propose GRUN, a general framework to enhances fine-tuning-based unlearning. GRUN leverages the shared mechanism between GA-based and suppression-based methods. It uses a soft gate function for distinguishing and a ReFT-based suppression module to adjust representations. GRUN improves both unlearning effectiveness and utility, and enables efficient unlearning.

\bibliography{custom}

\newpage

\appendix

\section{ReFT}
\label{appd:reft}

\begin{figure}[t]
\centering
  \includegraphics[width=\linewidth]{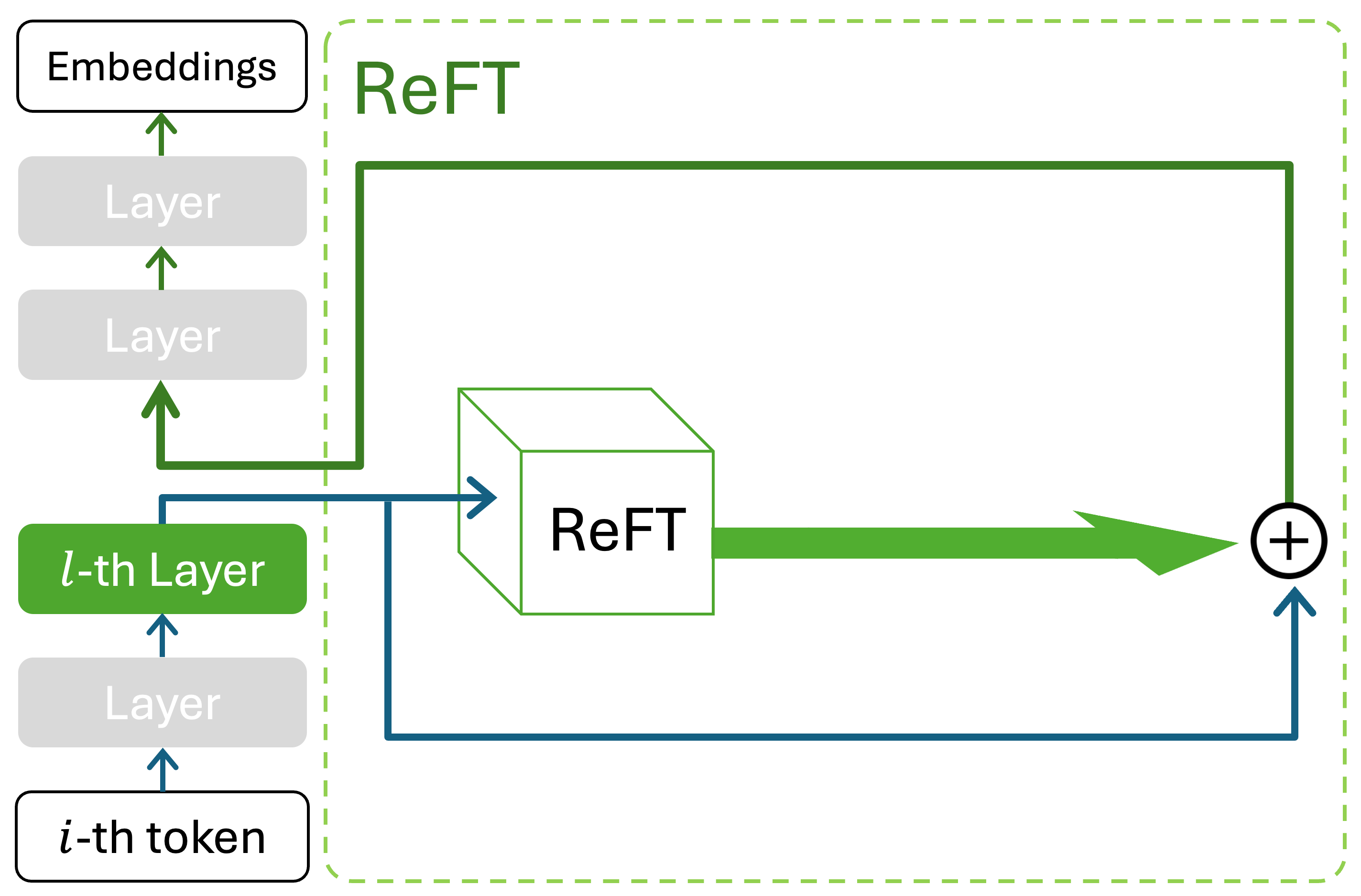}
  \caption{An overall of the framework ReFT. }
  \label{fig:reft}
  \vspace{-0.1in}
\end{figure}

The figure of ReFT is shown in Figure~\ref{fig:reft}.

\section{Hyper-parameters}

\subsection{Choosing layers for GRUN}
\label{appd:layer}

We find that when the layers are too close, it is possible to influence the each other's training. For example, when we use the last two layers for GRUN, the unlearning performance of Llama increases to 0.4 for GD. Thus, we use interval layers.

\subsection{The coefficient $c$ for sequential unlearning}
\label{appd:seq_c}

In our experiments, we tune the hyper-parameter $c$ to get the best utility while maintaining the unlearning. This is reasonable since the LLM builder have the target data which can be used to search the best hyper-parameters.

\section{Experimental settings.}
\label{appd:settings}

\subsection{Metrics}
\label{appd:metrics}

For Probability, TOFU uses the normalized likelihood for target and retaining data. For real authors and world facts, we follow their settings use the probability between correct answer and paraphrased answer (wrong answers). Please refer the details to \citet{maini2024tofu}.

\subsection{Other implementation details.}
\label{appd:details}

GRUN is trained for 40 epochs on NPO and IDK. All the learning rates are 1e-5. The time cost is tested on A6000 GPUs.

\section{Additional experiments}
\label{appd:g_output}

\begin{table}[t]
    \centering
    \resizebox{\linewidth}{!}{
    \begin{tabular}{cccccccc}
    \toprule
        \multirow{2}{*}{$L_{\text{G}}$} & \multirow{2}{*}{$L_{\text{u}}$} & \multicolumn{2}{c}{Gate 1 ($l=20$)} & \multicolumn{2}{c}{Gate 2 ($l=25$)} & \multicolumn{2}{c}{Gate 3 ($l=31$)} \\ 
        & & target $\uparrow$ & retain $\downarrow$ & target $\uparrow$ & retain $\downarrow$ & target $\uparrow$ & retain $\downarrow$ \\ \midrule
        \multirow{3}{*}{No} & GD &0.00&0.00& 0.99 & 0.08 & 1.00 & 0.05 \\ 
        & NPO & 1.00 & 1.00 & 1.00 & 1.00 & 1.00 & 1.00 \\ 
        & IDK & 1.00 & 1.00 &0.00&0.00&0.00&0.00\\ \midrule
        \multirow{3}{*}{Yes} & GD& 0.93 & 0.24 & 1.00 & 0.03 & 0.92 & 0.02 \\ 
        & NPO & 0.99 & 0.09 & 1.00 & 0.02 & 1.00 & 0.02 \\ 
        & IDK & 0.99 & 0.09 & 1.00 & 0.02 & 1.00 & 0.01 \\ \bottomrule
    \end{tabular}
    }
    \caption{Outputs of gate functions. $l=20, 25, 31$ represents the last 12th, last 7th and last layer respectively. The arrow $\uparrow$ (or $\downarrow$) means the output is expected to be close to 1 (or 0).}
  \label{tab:gate_output}
\end{table}

To further examine the different behaviors of the gate function with and without $L_{\text{G}}$, we present the gate function outputs for target and retaining data in Table~\ref{tab:gate_output}. 
With $L_{\text{G}}$, the gate function behaves as expected—opening for target data while closing for retaining data. Even in the absence of explicit guidance from $L_{\text{G}}$, the gate can still differentiate effectively, as seen in Gate 2 and Gate 3 of GD.  
For IDK, the gate function helps identify the optimal layer for ReFT and adjusts by closing redundant layers.
A special case arises with NPO when $L_{\text{G}}$ is absent: all gates remain open for both target and retaining data. Although this structure appears similar to ReFT-only, it has significantly enhanced unlearning effectiveness compared with ReFT-only. We conjecture that the soft gate influences the optimization process. In the case of ReFT-only, retaining data may compete with target data due to their reversed losses.  
For GRUN without $L_{\text{G}}$, the gate may prioritize forgetting data early in training, as the loss on retaining data has limited room to decrease—having already converged before unlearning. This hypothesis is supported by the observation that, within the first 10 steps, the forgetting loss of GRUN without $L_{\text{G}}$ is lower than that of ReFT-only.

\end{document}